\documentclass[10pt,twocolumn,letterpaper]{article}

\usepackage{iccv}
\usepackage{times}
\usepackage{epsfig}
\usepackage{graphicx}
\usepackage{amsmath}
\usepackage{amssymb}
\usepackage{url}
\graphicspath{ {./figures/} }

\usepackage[breaklinks=true,bookmarks=false]{hyperref}
\iccvfinalcopy 
\def\iccvPaperID{****} 

\ificcvfinal\fi

\begin{document}

\title{Perceptually Validated Precise Local Editing for Facial Action Units with StyleGAN}

\author{Alara Zindancıoğlu\\
Koç University\\
Istanbul, Turkey\\
{\tt\small azindancioglu15@ku.edu.tr}

\and
T. Metin Sezgin\\
Koç University\\
Istanbul, Turkey\\
{\tt\small mtsezgin@ku.edu.tr}
}

\maketitle

\begin{abstract}
  The ability to edit facial expressions has a wide range of applications in computer graphics. The ideal facial expression editing algorithm needs to satisfy two important criteria. First, it should allow precise and targeted editing of individual facial actions. Second, it should generate high fidelity outputs without artifacts. We build a solution based on StyleGAN~\cite{stylegan}, which has been used extensively for semantic manipulation of faces. 
  As we do so, we add to our understanding of how various semantic attributes are encoded in StyleGAN. In particular, we show that a naive strategy to perform editing in the latent space results in undesired coupling between certain action units, even if they are conceptually distinct. For example, although brow lowerer and lip tightener are distinct action units, they appear correlated in the training data. Hence, StyleGAN has difficulty in disentangling them. We allow disentangled editing of such action units by computing detached regions of influence for each action unit, and restrict editing to these regions. We validate the effectiveness of our local editing method through perception experiments conducted with 23 subjects. The results show that our method provides higher control over local editing and produces images with superior fidelity compared to the state-of-the-art methods. 

\end{abstract}

\section{Introduction}
\begin{figure*}
\vspace{-0.1in}
    \centering
    \includegraphics[width=0.9\linewidth]{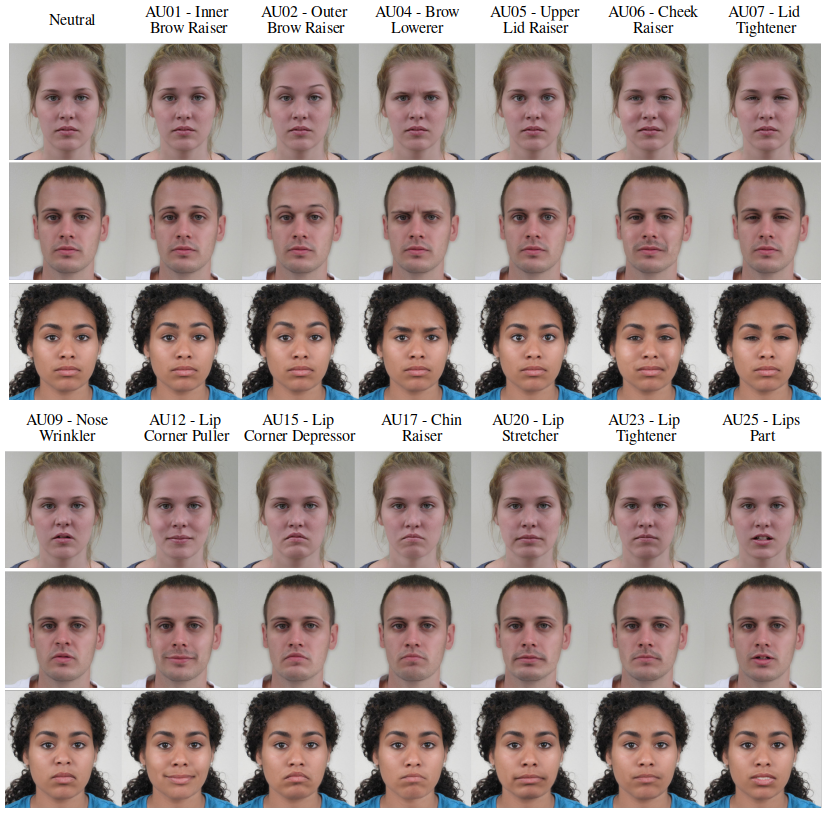}
   \caption{Sample editing results for individual action units.}
\label{fig:au_edits}
\vspace{-0.1in}
\end{figure*}
Facial expression editing or synthesis aims to manipulate the expression of a given face image, while preserving the personal identity and other facial features. Due to the complex structure of human face, the task is especially challenging since it involves synthesizing unseen muscle movements realistically. Photorealistic facial expression editing has wide range of applications in various fields such as human-computer interaction, entertainment industry, virtual reality, and photography.
. 

The task of facial expression editing has been approached with traditional computer vision, as well as deep learning based methods. Traditional approaches mainly make use of mapping, warping, and morphing operations~\cite{face_decomposition, polynomial_maps, expression_mapping, zhang2005geometry, yang2011expression}. These approaches require excessive supervision of operations such as extracting complex geometric information from the face, computing flow maps, or doing face decomposition. Approaches based on deep learning attempt to build end-to-end generative models to synthesize face images with multiple facial expressions~\cite{ding2018exprgan, choi2018stargan, song2018geometry, qiao2018geometry, ganimation}. With the advancements in variational autoencoders (VAEs) and generative adversarial networks (GANs), impressive results have been reported in face synthesis, facial expression editing, and transfer. The main procedure with generative models is to encode images into a semantically meaningful latent space and perform editing by manipulating the latent vector. 

The field of facial expression editing is closely related to Psychology. Facial Action Coding System (FACS) developed by Paul Ekman~\cite{facs} is widely used in the area to label facial actions. It describes emotions and related facial movements in terms of Action Units (AUs). In some works, AU activations are used as labels or conditions for face editing with generative models~\cite{deep_belief_nets, ganimation}.

With successful GAN architectures capable of synthesizing photorealistic face images, a recent research approach has been to utilize pre-trained GAN models for image editing. Along the same line, we build our solution for facial action unit editing on the state-of-the-art StyleGAN model.

Understanding how various semantic attributes are encoded within the latent space of StyleGAN serves as a powerful means for image manipulation.
 
The mechanism through which the latent space of StyleGAN maps to various semantic attributes is not well understood, but there have been notable attempts to characterize it. Some authors have proposed
finding linear directions corresponding to specific attributes within the StyleGAN latent space. For example, Shen  et al.~\cite{shen2020interpreting} show that there are linear directions within the latent space corresponding to certain attributes such as pose, age, gender, and the presence of a smile. Others have identified certain layers and channels within the StyleGAN architecture that control discrete semantic attributes~\cite{ganlocalediting}. Unfortunately, neither of these approaches successfully characterize variations in facial action units which are localized, fine scale and delicate in nature. We add to our understanding of the latent space of StyleGAN by showing that fine scale variations in facial action units require manipulating the latent vector representation of a source content, as well as identifying and adjusting localized patches within the StyleGAN layers and activation maps. To achieve this we present two complementary methods: first is global editing, where latent encodings of images are manipulated in a way that does not necessarily focus on specific regions on the resulting image, and local editing, which refers to restricting the global edit only to a region of interest. Sample editing results are shown in Figure~\ref{fig:au_edits}.

\begin{figure*}
    \centering
    \includegraphics[width=1\linewidth]{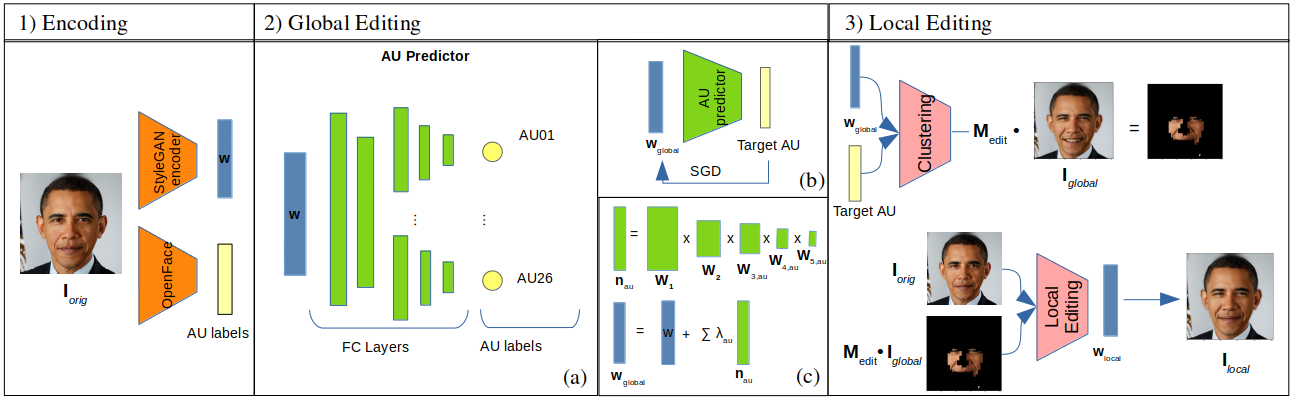}
   \caption{Our pipeline for image editing in the latent space of StyleGAN. 1) Images are first encoded in the latent space of StyleGAN and AU labels are prepared with OpenFace~\cite{openface0}. 2) Latent vectors are fed to the global editor, which is essentially an AU predictor (a) utilized in two different ways for editing: (b) via optimization and (c) by extracting linear directions. 3) Finally local editing is performed on globally edited latent vector $\mathbf{w}_{global}$.}
\label{fig:pipeline}
\vspace{-0.1in}
\end{figure*}

Our contributions are as follows: 
\vspace{-0.1in}
\begin{itemize}
  \item We present a simple yet effective way of facial action unit editing in the latent space of StyleGAN, producing high-fidelity editing results. 
  \vspace{-0.1in}
  \item We offer methods for both global and local editing of facial action units. Our approach allows adjusting the strength of the action units on a continuous scale and joint editing of multiple action units. 
  \vspace{-0.1in}
  \item We add to our understanding of the latent and activation spaces of StyleGAN by analysing how semantics of facial action units are encoded.
  \vspace{-0.1in}
  \item We show that combining editing techniques in latent and activation spaces improves the editing quality for facial action units.

\end{itemize}

\section{Related Work}
Facial expressions are essential in social communication since they convey important information about people's internal emotional states. Facial Action Coding System (FACS) represents facial muscle movements that are related to emotions through Action Units (AUs). There are a number of AUs defined and their combinations produce a wide range of facial expressions and emotions. AU recognition task has long been studied, and it is usually combined with facial landmark and emotion recognition tasks~\cite{tadas_au_detection, ruiz2015emotions, onal2019d}. 
\subsection{Deep generative models}
Common deep generative models for image synthesis are variational autoencoders (VAEs) and generative adversarial networks (GANs). VAEs are probabilistic models with an encoder-decoder architecture. GAN models are composed of a generator and a discriminator trained sjointly in a minimax game setup. With recent improvements in their architectures and loss functions~\cite{liu2017unsupervised, kingma2013auto, larsen2016autoencoding}, we often see combined VAE-GAN architectures to boost performance and functionalities. In addition to realistic image synthesis, deep generative models are widely used for a wide range of tasks such as conditional image generation~\cite{mirza2014conditional}, image-to-image translation~\cite{isola2017image,liu2017unsupervised, choi2018stargan}, super-resolution imaging~\cite{ledig2017photo}, and facial attribute editing~\cite{perarnau2016invertible, shen2020interpreting, he2019attgan, liu2019stgan}.

\subsection{Facial expression editing}
Facial expression editing is an intricate task as it requires manipulation of various face regions simultaneously and realistically while maintaining the identity of the face. Some generative models are conditioned on discrete emotion categories~\cite{choi2018stargan, ding2018exprgan, zhou2017photorealistic}. StarGAN~\cite{choi2018stargan} uses a single model for multi-domain image translation, which can be utilized for facial expression transfer. Others use facial geometry in order to perform continuous editing of facial expressions on a continuous range of intensities~\cite{qiao2018geometry, song2018geometry, ganimation}. Facial expression transfer is another problem that has been widely studied, where the aim is to transfer the expression of one face to another~\cite{xu2017face,wiles2018x2face}.

There are also lines of work that utilize action unit labels to perform editing on the face ~\cite{tripathy2020icface, ganimation, wu2020cascade}. All aformentioned methods, except~\cite{ganimation, wu2020cascade} are methods that process the image as a whole. GANimation~\cite{ganimation}, on the other hand, incorporates an attention later that focuses the network on regions of the face that are most relevant to synthesize the target expression. Another work that focuses on local regions is Cascade EF-GAN~\cite{wu2020cascade}, which processes images via global and local branches that focus on selected expression-intensive regions - eyes, nose, and mouth.

\subsection{Utilizing pre-trained models}
Utilizing pre-trained GAN models for image editing has attracted attention recently. Since most successful GANs for image synthesis generate images based on a random latent code, real images first need to be encoded in the latent space of the specific GAN for subsequent use. There are several GAN-inversion techniques that allow encoding real images in the latent space. While some methods train and encoder neural network to find the latent representation ~\cite{chai2021using, kingma2013auto}, others optimize the latent code such that the generated image matches the target image~\cite{stylegan-encoder, creswell2018inverting, abdal2019image2stylegan}.

Two main approaches to manipulate images with a pre-trained GAN are investigating the semantics in the latent space to perform edits on the latent vectors and manipulating the activation maps. Latent code based methods make use of intrinsic semantics of the latent space in order to learn a manifold or find directions through which latent vectors can be moved~\cite{shen2020interpreting, nitzan2020disentangling}. 

Activation based methods focus on the semantics learned by feature maps of convolutional layers. While some manipulate spatial positions on the activation tensors of certain layers~\cite{bau2018gan,suzuki2018spatially}, others study if certain feature maps focus on specific objects and perform editing by only manipulating those channels~\cite{ganlocalediting}. 

Our work can be viewed as a combination of latent code and activation based methods. In order to perform global edits on the input image, we use latent space manipulation and investigate the disentangled semantics learned in the StyleGAN latent space. We also present two techniques for local editing: optimizing the latent encoding and patching the activation maps. In addition to editing, we study the semantics learned by the latent and activation spaces of StyleGAN.  

\begin{figure}
    \centering
    \includegraphics[width=1\linewidth]{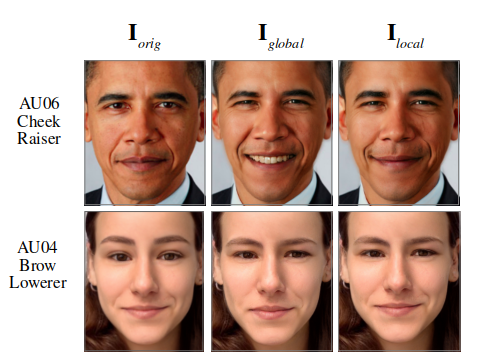}
   \caption{With global editing of some action units, changes at unrelated regions of the face can be observed ($\mathbf{I}_{global}$). Local editing restricts  manipulations to corresponding regions ($\mathbf{I}_{local}$). }
\label{fig:global_local}
\vspace{-0.1in}
\end{figure}

\section{Methodology}
Instead of building a generative adversarial network from scratch, we use the state-of-the-art GAN model StyleGAN to synthesize images, and perform editing in its latent and activation spaces. Given an input image $\mathbf{I}_{orig}$, we first encode it to the intermediate latent space $\mathcal{W}$ of StyleGAN using StyleGAN-encoder~\cite{stylegan-encoder} to obtain $\mathbf{w}_{orig}$. Later, $\mathbf{w}_{orig}$ is fed to the \textit{global editor network}, which is essentially an action unit detector but is utilized to edit latent vectors with respect to desired action unit intensities. The details of global editing are discussed in Section 3.1. As we elaborate later, our experiments show that some action units appear to be entangled with each other, i.e. changing certain action units cause changes at parts of the face that were not intended to change. To overcome this problem, we use \textit{spherical k-means clustering}~\cite{buchta2012spherical, ganlocalediting} to find local regions of the face that correspond to each action unit. Together with these local regions, globally edited latent vector $\mathbf{w}_{global}$ is fed to the \textit{local editor network}, which modifies only the desired local regions of the face. The overall model architecture is demonstrated in Figure~\ref{fig:pipeline}.

We created an encoded dataset consisting of 10,323 images from the RaFD~\cite{rafd}, Bosphorus~\cite{bosphorus}, and CFEE~\cite{cfee} datasets. For labeling, we use the OpenFace software~\cite{openface0, openface1} and get regression values of 16 action units for each image. This datasets is used to train the global editor network.

\subsection{Global Editing}
For global editing, we train an action unit detector that predicts action unit values from the latent encodings of images. The network takes as input the latent vectors $\mathbf{w}$ and outputs a vector of regression values for 16 action units $\mathbf{AU}$. The AU detector is a feedforward network with two common fully connected layers followed by 16 AU branches, each with three fully connected layers. ReLU activation function is used at each layer except the output nodes, where linear activation function is used. The network is trained with MSE loss and SGD optimizer. 

There are two ways in which we use this network to edit action units in the latent space. The first one is an optimization based approach, where we update the $\mathbf{w}$ vector via gradient descent by setting the target action unit values as the output of the AU detector. This approach allows precise editing by only changing specific action unit values and keeping others the same. It can also be used to transfer the target expression from one image to another. But this approach is slow due to its iterative nature. 

A faster approach is to get weights of the network after training, and take the matrix multiplication of the weights to get a linear direction in the latent space for each action unit. Let $\mathbf{W}_{1}$ and $\mathbf{W}_{2}$  refer to weight matrices of first two layers of the global editor network, and $\mathbf{W}_{3,au}$, $\mathbf{W}_{4,au}$, and $\mathbf{W}_{5,au}$ refer to the third, forth, and fifth layers of network branches that correspond to the action unit $au\in\textbf{AU}$. The linear direction $\mathbf{n}_{au}$ in the latent space for editing action unit $au$ is then given by,
\begin{equation}
    \mathbf{n}_{au} = \mathbf{W}_1 \cdot \mathbf{W}_2 \cdot \mathbf{W}_{3,au} \cdot \mathbf{W}_{4,au} \cdot \mathbf{W}_{5,au}
\end{equation}
To edit multiple action units at once, we take linear combinations of them with varying intensities. Let $\lambda_{au}$ be the intensity of $au$ and \textbf{AU}\textsubscript{edit} be the list of action units to be edited. Combined direction \textbf{n}\textsubscript{edit} is found by
\begin{equation}
     \mathbf{n}_{edit} = \mathbf{W}_1 \cdot \mathbf{W}_2 \cdot\lambda_{au}\sum\limits_{\mathbf{AU}_{edit}}{\mathbf{W}_{3,au} \cdot \mathbf{W}_{4,au} \cdot \mathbf{W}_{5,au}}
\end{equation}
Adding the resulting vector to the $\mathbf{w}_{orig}$ moves it in a linear direction that enhances the desired action unit value. In both approaches, optimization and moving in linear directions, multiple action units can be modified with varying intensities. 

With global editing, especially in the fast approach, entanglement between some action units that often occur together is observed. Figure~\ref{fig:global_local} shows the result of entanglement as a result of modifying certain action units. This can be explained by dataset or latent space bias. To suppress editing in regions where changes were not intended, we further optimize the $\mathbf{w}$ vector with local editing. 
\subsection{Local Editing}

\begin{figure}
\hspace{-0.2in}
    \centering
    \includegraphics[width=1\linewidth]{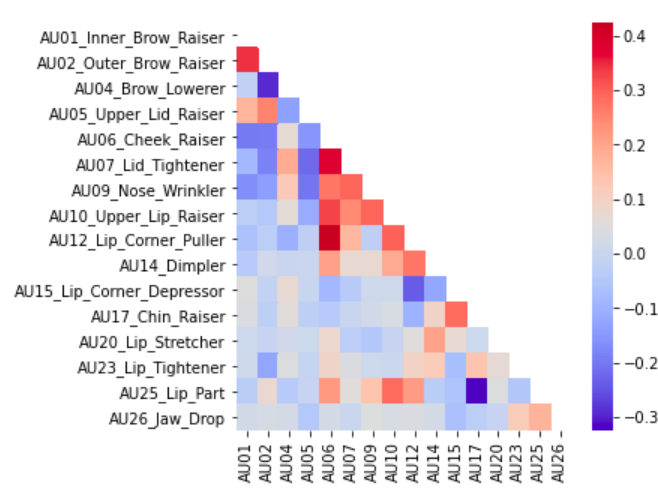}
   \caption{Correlation between action units.}
\label{fig:correlation}
\vspace{-0.15in}
\end{figure}

For local editing, we first find semantic regions on the face by applying spherical k-means clustering on the activation maps. We then propose two ways of performing local edits on the face: first is optimizing the latent encoding, and second is multiplexing activation maps. 
\subsubsection{Spherical K-means Clustering}
Following Collins \textit{et al.}~\cite{ganlocalediting}, we apply spherical k-means clustering to find local regions learned in the StyleGAN activation space. Clustering is applied to the 7th layer activation maps $\mathbf{A} \in \mathbb{R}^{N\times C\times H\times W}$ 
where N is the number of images, C is the number of channels, and H and W are height and width. With K clusters, we obtain a cluster catalog which outputs a tensor of cluster memberships, $\mathbf{U} \in \{0,1\}^{N\times K\times H\times W}$.
From K clusters, only those that fall within the face region are used. 
Using encoded vectors of images from RaFD~\cite{rafd}, Bosphorus~\cite{bosphorus} and CFEE~\cite{cfee} datasets, we train the clustering network and determine regions on the face that are most affected by global editing of individual action units. Details of this analysis are given in section 4.2. With both visual inspection of facial regions and the results of this study, we construct a dictionary of action unit-cluster relationships. Thus for each action unit, we have a list of associated clusters.

\begin{figure}
    \centering
    \includegraphics[width=1\linewidth]{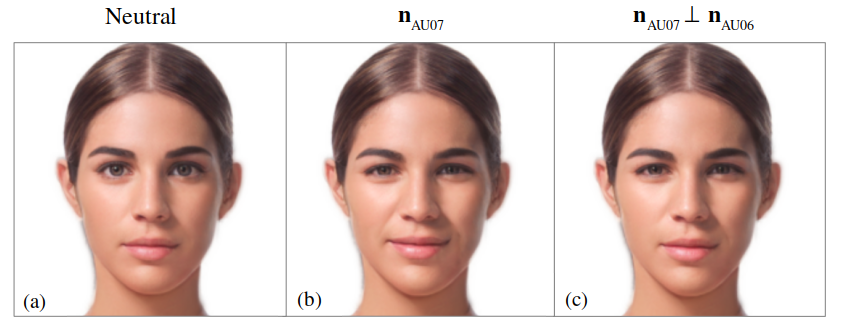}
   \caption{Moving in orthogonal directions. Linear direction vector $\mathbf{n}_\mathrm{AU07}$ for AU07 - Lid Tightener is highly correlated with direction vector $\mathbf{n}_\mathrm{AU06}$ for AU06 - Cheek Raiser (b). Moving in the direction $\mathbf{n}_\mathrm{AU07}\perp\mathbf{n}_\mathrm{AU06}$ results in only editing AU07 without effecting AU06 (c).}
\label{fig:orthonormal}
\vspace{-0.1in}
\end{figure}
\subsubsection{Optimizing the latent encoding}
Let $\mathbf{w}_{global}$ be the latent vector of the globally edited image, and \textbf{AU}\textsubscript{edit} be the list of action units to be edited. The latent vector is first fed to the StyleGAN generator to compute the 7th layer activation maps $\mathbf{A} \in \mathbb{R}^{512\times32\times32}$ 
and the output image
$\mathbf{I}_{global} \in \mathbb{R}^{1024\times1024\times3} $. Then, spherical k-means clustering is applied on the activations $\mathbf{A}$ to obtain the local clusters $\mathbf{U} \in \{0,1\}^{K\times32\times32}$. The local clusters are resized to the output image size and only k clusters that are related to \textbf{AU}\textsubscript{edit} are selected to obtain the tensor $\mathbf{U}\in\{0,1\}^{k\times1024\times1024}$. Note that clusters are selected from a pre-computed dictionary of AU-cluster relationships (Section 4.2). The mask for local editing  $\mathbf{M}_{edit}$ is calculated by,
\begin{equation}
    \mathbf{M}_{edit} = \sum_{k}\mathbf{U}_{k,1024,1024}
\end{equation}
To locally edit the image, we update the latent vector $\mathbf{w}_{global}$ by minimizing the L2 loss between the output image $\mathbf{I}_{local}$ and globally edited image $\mathbf{I}_{global}$ within the editing mask region. Editing loss is defined as:
\begin{equation}
    \mathcal{L}_{edit} = \frac{1}{N} \sum_{n=1}^{N}(\mathbf{M}_{edit} \odot (\mathbf{I}_{local} - \mathbf{I}_{global}))^{2}
\end{equation}
$\odot$ denotes elementwise multiplication. 

Outside the editing mask, we minimize the L2 loss between the output image $\mathbf{I}_{local}$ and the original image $\mathbf{I}_{orig}$. Regions outside editing mask are defined as $1-\mathbf{M}_{edit}$. Consistency loss is defined as:
\begin{equation}
\mathcal{L}_{const} = \frac{1}{N} \sum_{n=1}^{N}((1-\mathbf{M}_{edit}) \odot (\mathbf{I}_{local} -\mathbf{I}_{orig}))^{2}
\end{equation}
Finally, the total loss is given by,
\begin{equation}
\mathcal{L} = \alpha \mathcal{L}_{edit} + \beta \mathcal{L}_{cons}
\end{equation}
After the optimization, the locally edited latent vector $\mathbf{w}_{local}$ is obtained. Finally the locally edited image $\mathbf{I}_{local}$ can be generated by feeding the vector to StyleGAN.

\subsubsection{Multiplexing Activation Maps}
Another way of performing local editing is multiplexing activation maps, which is faster but does not yield the locally edited latent vector $\mathbf{w}_{local}$. As in optimization based local editing, first $\mathbf{M}_{edit}$ is obtained by applying spherical k-means clustering on the activation maps of globally edited latent vector $\mathbf{w}_{global}$. This time the mask is not resized to the output image size, as it will be applied on he activation maps of size (512, 32, 32). As the original latent vector $\mathbf{w}_{orig}$ is sent through the network, the patch of 7th layer activation maps $\mathbf{A}_{orig}$ that corresponds to the \textbf{AU}\textsubscript{edit} is replaced by the patch of activation maps from the globally edited image $\mathbf{A}_{global}$.  
\begin{equation}
    \mathbf{A}_{local} = (1-\mathbf{M}_{edit}) \odot \mathbf{A}_{orig} + \mathbf{M}_{edit} \odot \mathbf{A}_{global}
\end{equation}
With the locally edited activation map $\mathbf{A}_{local}$, StyleGAN generator successfully incorporates the patch and outputs the locally edited image. Resulting image does not contain artifacts from the patching, since the operation is performed in the intermediate layers and the generator blends the patch as it passes through the subsequent layers. 

\section{Experiments}
In this section, we analyze the latent and activation space of StyleGAN, and report valuable insights on the semantics of the activation maps. We also compare our  editing results with the state-of-the-art models for facial expression editing.

\subsection{Linear Directions and Entanglement}
There are multiple ways of finding linear directions in the latent space of StyleGAN. For example Shen \textit{et al.} train independent SVMs on pose, smile, age, gender, and eyeglasses attributes\cite{shen2020interpreting} to find linear hyperplanes that separate the latent space such that samples that possess an attribute are placed on the same side. Instead of training independent SVMs for each action unit, we train one network with multiple branches and shared weights, which we refer as the \textit{the global editor network}. We obtain linear directions by multiplying the weights of this network (Section 3.1). Moving the latent vector in the positive direction that corresponds to an action unit enhances the action unit value. Similarly, moving in the negative direction reduces the action unit intensity. 

Although action units are conceptually distinct, some are activated together in most datasets that are available. For example, brow lowerer and lip tightener, or cheek raiser and lips part action units are distinct and can be performed individually, but they are commonly observed together in emotion related datasets. Thus some of the linear directions that are found are not totally disentangled with the others. To demonstrate this, we find the correlation of each action unit with respect to the others by calculating the size of projected vector. The correlation graph is presented in Figure~\ref{fig:correlation}. Orthogonal portions of linear direction vectors can be extracted to decouple editing between correlated action units. Figure~\ref{fig:orthonormal} shows the result of this operation.

\vspace{-0.05in}
\subsection{Insights on the Activation Maps}

\begin{table}
\vspace{-0.1in}
\hspace{-0.2in}
    \centering
    \includegraphics[width=1.05\linewidth]{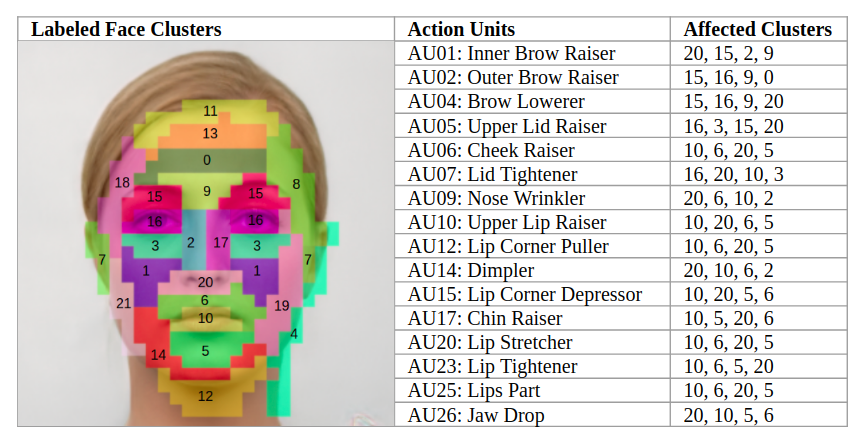}
   \caption{Face clusters obtained by spherical k-means clustering. Affected clusters column presents top four relevant clusters for each action unit computed with Equation~\ref{eq:au_cluster}.}
\label{fig:clustering}
\vspace{-0.2in}
\end{table}

For local editing, we need to form a dictionary of AU-cluster relationships so that editing is limited only to regions that are relevant. To do so, we compare channel activations of neutral and edited images at the clustering regions. This also gives us an insight on how well the global editor network localizes. 

After visual inspection and consulting previous research~\cite{ganlocalediting, stylegan}, we decide that 7th layer with $32\times32$ resolution is the best layer for semantic facial part learning. Using the global editor network, we first edit 180 neutral images from the RaFD dataset for each action unit. Then we feed both edited and neutral vectors to StyleGAN in order to get 7th layer activation maps. We obtain the resulting tensor $\mathbf{A}\in\mathbb{R}^{AU\times N\times C\times H\times W}$, where $AU$ denotes action unit index, $N$ is number of neutral images, $C$ is number of channels, and $H$ and $W$ are spatial dimensions of the layer. In our experiments, $(AU, N, C, H, W) = (17, 180, 512, 32, 32)$, where $AU=0$ denotes the neutral image and $AU\in[0,16]$ are action unit indices. 

For each neutral and edited image, we also the compute local clusters with spherical k-means clustering to obtain the tensor $\mathbf{U} \in \{0,1\}^{AU\times N\times K\times H\times W}$, where K denotes the number of clusters.
\begin{figure}
\hspace{-0.2in}
    \centering
    \includegraphics[width=1.05\linewidth]{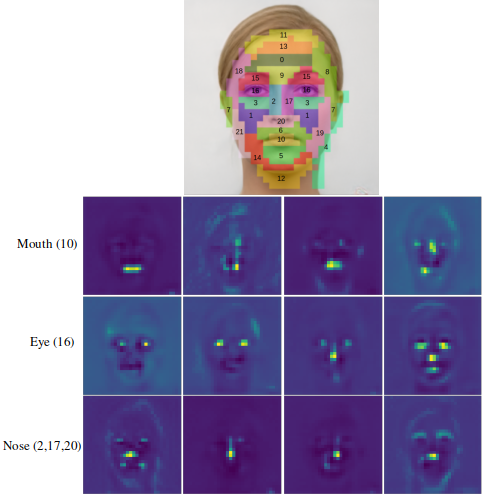}
   \caption{Four most relevant channel outputs for the mouth, eye, and nose regions.}
\vspace{-0.1in}
\label{fig:activations}
\end{figure}
Using these matrices, our aim is to find: 1) regions that are most affected by global editing of action units and 2) channels that are most activated by certain local regions on the face. With the following operation, we obtain the AU-cluster-channel relationship matrix $\mathbf{R}\in\mathbb{R}^{AU, K, C}$:
\begin{equation}
\mathbf{R}_{au,k,c} = \sum\limits_{n}(\frac{\sum\limits_{h,w}(\mathbf{U}
_{au,n,k,h,w} \odot (\mathbf{A}_{au,n,c,h,w} - \mathbf{A}_{0,n,c,h,w})^{2})}{\sum\limits_{h,w} \mathbf{U}_{au,n,k,h,w}})
\label{eq:au_k_c}
\end{equation}
The AU-cluster relationship is computed by:
\begin{equation}
\mathbf{R}_{au,k} = \sum\limits_{c} \mathbf{R}_{au,k,c}
\label{eq:au_cluster}
\end{equation}
Table~\ref{fig:clustering} presents most affected facial regions by each action unit via global editing. It can be seen from the table that global editing affects irrelevant parts of the face for some action units. For example, editing Inner Brow Raiser affects the nose area (cluster 20) the most, which is not expected. The reason for this may be the bias in the datasets used for training the global editor network, or latent space bias in StyleGAN. 

\begin{figure}
\hspace{-0.2in}
    \centering
    \includegraphics[width=1.05\linewidth]{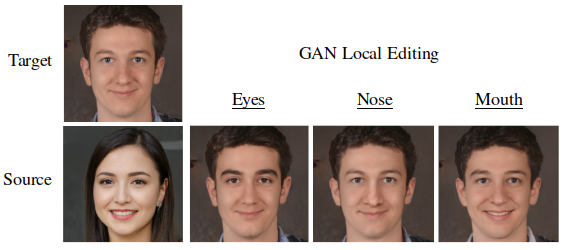}
   \caption{Results by GAN Local Editing ~\cite{ganlocalediting}}
\label{fig:ganlocalediting}
\vspace{-0.1in}
\end{figure}

Furthermore, we inspect the facial region and channel relationship in the StyleGAN activation space, independent of action units. Previous research by Collins \textit{et al.} ~\cite{ganlocalediting} suggests that certain channels of the convolutional layers of StyleGAN focus on specific regions. Following their approach we compute facial region - channel relationship with a slight modification on Equation~\ref{eq:au_k_c}, 
\begin{equation}
\mathbf{R}_{k,c} = \sum\limits_{n}(\frac{\sum\limits_{h,w}(\mathbf{U}
_{n,k,h,w} \odot (\mathbf{A}_{n,c,h,w})^{2})}{\sum\limits_{h,w} \mathbf{U}_{n,k,h,w}})
\label{eq:cluster_activmap}
\end{equation}

\noindent which outputs the contribution of each channel towards each semantic cluster. Collins \textit{et al.} perform local editing by interpolating activation maps of the source and target images at a certain layer with slope of interpolation defined by the tensor of region - channel relationships $\mathbf{R}_{k,c}$. Assigning  more weight to the relevant channels that affect a region results in transferring local attributes from source image to the target image~\cite{ganlocalediting}. To demonstrate localization of convolutional channels, we visualize activation maps that are most relevant to three regions: mouth, eyes and nose in Figure~\ref{fig:activations}. Just by visual inspection, it can be seen that even the most responsible channels for a region have high activation on parts of the face that are not relevant. Thus performing local editing with this approach results in either editing parts of the face that were not intended, or very weak editing results where the change is hard to observe. Editing results by Collins \textit{et al.} (GAN Local Editing) are presented in Figure~\ref{fig:ganlocalediting}, where we can observe a loss in the target image's identity because of unlocalized attribute transfer. 

Thus when we perform local editing via manipulation of activation maps, we do not interpolate activation maps as a whole, but transfer the patch of corresponding region at every channel in order to avoid transfer of attributes outside the region of interest or loss of intensity of attributes at a local region. The difference between two methods are presented at Section 5. 
\begin{figure*}
    \centering
    \includegraphics[width=1.05\linewidth]{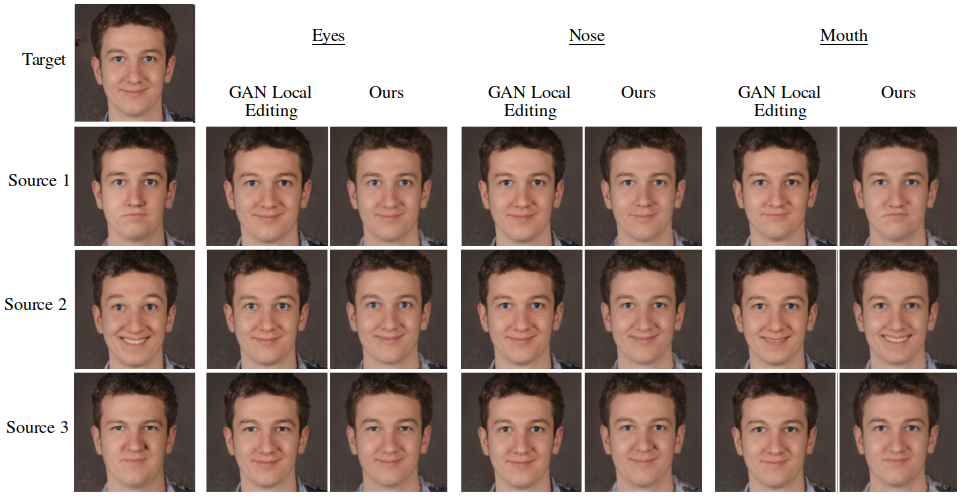}
    
   \caption{Comparison of local editing results. We transfer attributes from the local regions eyes, nose and mouth of source images to the target image using GAN Local Editing~\cite{ganlocalediting} and our local editing method. Especially in nose and mouth regions, it can be seen that GAN Local Editing cannot fully transfer the attributes. For example, the mouth region of Source 2 is poorly transferred with GAN Local Editing, while our approach transfers the same mouth to the target without affecting other regions on the face.}
\label{fig:comparison_ganlocalediting}
\vspace{-0.1in}
\end{figure*}

\begin{figure}
\hspace{-0.2in}
    \centering
    \includegraphics[width=1.05\linewidth]{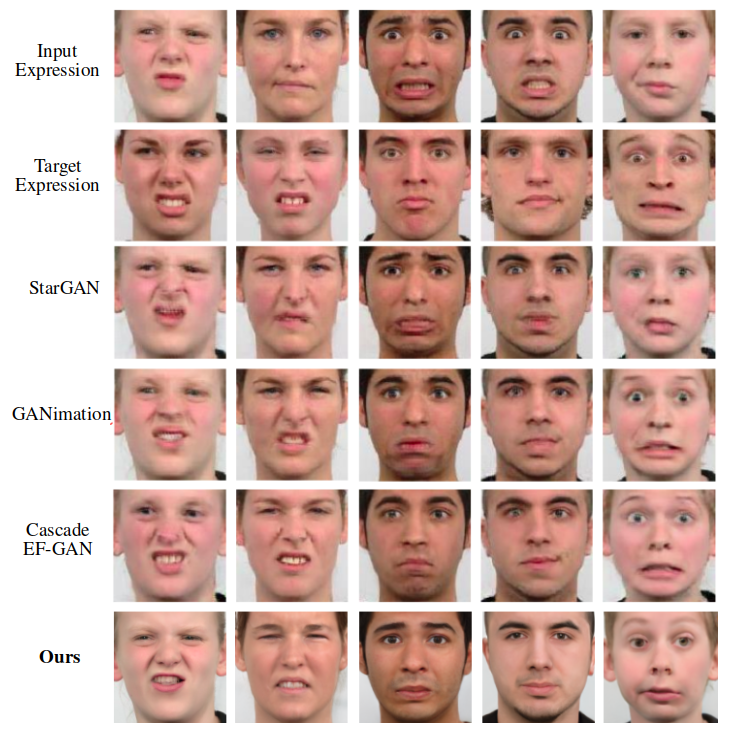}
       \caption{Comparison of facial expression transfer results. According to the user study (Figure \ref{fig:user_study}), our method produces much less artifacts and preserves the identity of the input image better. Our method also outperforms others, except Cascade EF-GAN, in expression transfer. }
    \vspace{-0.2in}
    \label{fig:comparison_expression}
    \vspace{-0.2in}
\end{figure}
\begin{figure}
    \centering
    \includegraphics[width=0.7\linewidth]{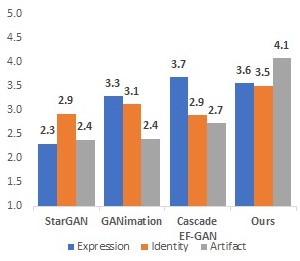}
   \caption{T-tests show that the perceived amount of artifacts is significantly less in our model compared to others (p$<$0.001). Identity preservation is also rated better (p$<$0.05). For expression transfer, our model performs significantly better than StarGAN (p$<$0.01) and GANimation (p=0.11). Although Cascade EF-GAN performs better in expression transfer, the difference is not significant (p=0.43).}
\label{fig:user_study}
\vspace{-0.1in}
\end{figure}
\begin{figure*}
    \centering
    \includegraphics[width=0.9\linewidth]{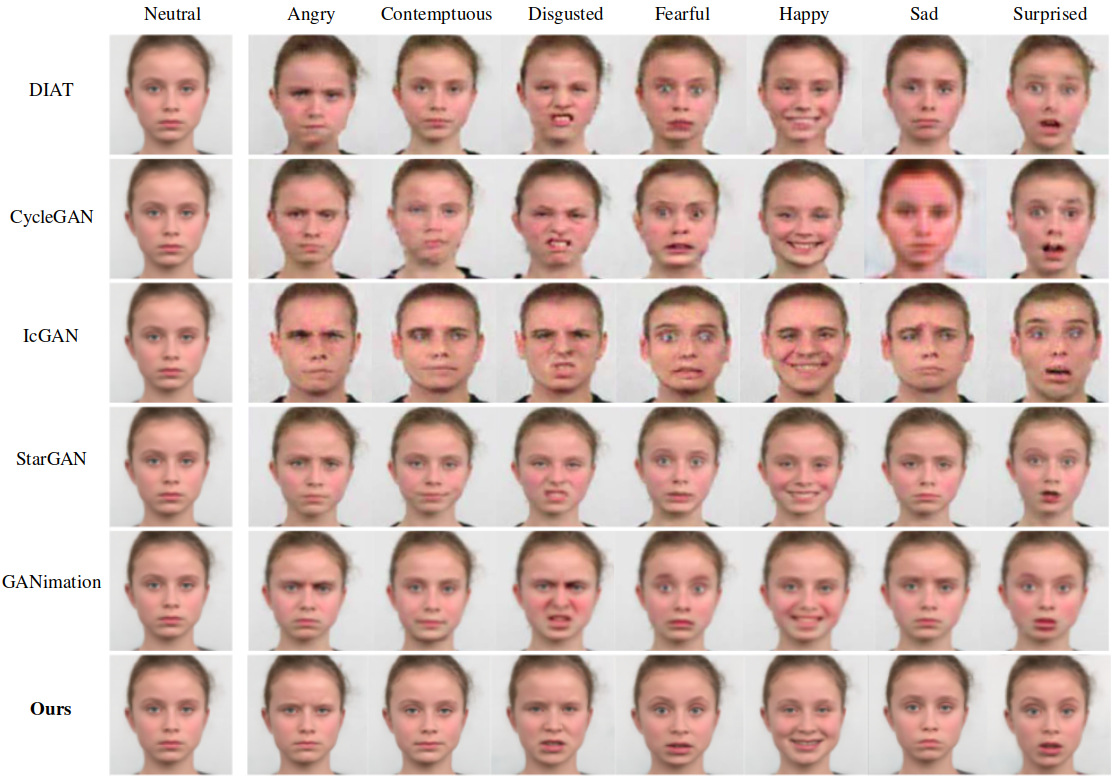}
   \caption{Comparison results on emotion editing. Our method produces much less artifacts than others and successfully synthesizes the correct emotion while preserving the identity.}
\label{fig:comparison_emotion}
\vspace{-0.1in}
\end{figure*}

\section{Comparisons and User Study}
We compare our editing results with the state-of-the-art facial expression editing models for three main tasks: local editing, facial expression transfer and emotion editing.

For local editing in the activation space, we compare our results with GAN Local Editing by Collins \textit{et al.}~\cite{ganlocalediting}. We use the same technique to find local regions on the face by applying spherical k-means clustering on the activation maps of a certain layer. After determining the regions, they compute the contribution of each channel to the region of interest and interpolate activation maps by giving more weight to the relevant ones. As discussed in Section 4.2, we show that most activation maps do not focus on specific semantic regions, but influence multiple locations on the face. It can be seen in Figure~\ref{fig:activations} that even the most relevant channels for a region affect irrelevant parts of the face. So transferring attributes by selecting activation maps based on their contribution to a region and discarding seemingly unrelated activation maps result in either transferring attributes from outside the region of interest or losing the intensity of attributes in the region. Based on our analysis of the activation space, we suggest that channels need to be considered cumulatively in StyleGAN. Thus we transfer local patches cut out from all activation maps of a layer from the source image to the target image. Comparison results are presented in Figure~\ref{fig:comparison_ganlocalediting}. It should be noted that unlike GAN Local Editing, our purpose is to transfer local attributes between original and edited images of the same person. Using our method for transferring local attributes between different identities would produce artifacts because of differences such as dimensions of the face or skin color. In order to transfer attributes between different identities, we would suggest using a combination of both approaches, i.e. interpolating activation maps and patching the region of interest, so that undesired changes do not appear at irrelevant parts of the face as in Figure~\ref{fig:ganlocalediting}.

For facial expression transfer, we first predict action unit values of the target image using the action unit predictor. Then we update the latent vector corresponding to the input expression via gradient descent to match the target expression. Results can be seen in Figure~\ref{fig:comparison_expression}. We performed a user study with 23 participants to compare our editing results with StarGAN~\cite{choi2018stargan}, GANimation~\cite{ganimation} and Cascade EF-GAN~\cite{wu2020cascade} models on facial expression transfer. We asked participants to evaluate edited images on three conditions: success in facial expression transfer, preservation of identity, and lack of artifacts. Resulting ratings on Likert scale are provided in Figure~\ref{fig:user_study}. The results show that our method outperforms others significantly for absence of artifacts on the edited image and preservation of identity from the input image. For expression transfer, our method performs significantly better than StarGAN and GANimation. Although Cascade EF-GAN is rated more successful for expression transfer, T-test shows that the difference is not significant (p=0.43).

To reproduce the emotions, we refer to studies on FACS by Paul Ekman~\cite{ekman1997face} to understand which action units should be activated for each emotion. We then perform combined editing of action units by boosting the intensity of the relevant ones. Comparison results with the models DIAT~\cite{li2016deep}, CycleGAN~\cite{zhu2017unpaired}, IcGAN~\cite{perarnau2016invertible}, StarGAN and GANimation are shown in Figure~\ref{fig:comparison_emotion}. It should be noted that although our method is not originally designed for expression transfer or emotion editing, it allows us to manipulate multiple action units at once with varying intensities to perform high-fidelity expression editing.

\section{Limitations and Societal Impact}
Although using the pre-trained StyleGAN model has many advantages, such as access to a rich latent space and impressive generative power, it has limitations when working with real images. First, in order to work with real images their latent representations must be obtained. The best methods for latent encoding are iterative optimization-based approaches, which is very time consuming. Second, images produced by the encoded representations are not exactly the same with the original image. It can be observed that encoded facial images preserve identity but have a smoother skin, less wrinkles and facial marks. In our research, we paid special attention to identity preservation while image encoding and action unit transfer, but preservation of finer details of the original face image is still a challenge. 

The ability to control action units with the high degree of fidelity and precision afforded by our algorithm can have malicious applications in the form of deep fakes. However, it also opens up avenues for remarkable  applications. In fact, the original motivator for our entire work has been the prospects of using it for rehabilitation of individuals suffering from Autism Spectrum Conditions, particularly those who have difficulty in producing facial expressions corresponding to specific emotional displays. Formative feedback in the form of an ASC sufferer's own face correctly displaying emotions pushes the boundaries of formative feedback-based rehabilitation systems beyond the current state of the art~\cite{Schuller2015}.

\section{Conclusion}
In this paper we present methods for both global and local editing of facial images by utilizing the state-of-the-art StyleGAN model. The idea can be extended to any domain of editing or GAN architecture, with suitable labeling of datasets. Leveraging the latent properties and generative power of a pre-trained GAN architecture allows us to produce high-fidelity editing results without the burden of training a GAN from scratch. Our analysis of the latent and activation spaces of StyleGAN also provides valuable insights that add to our understanding of how action unit semantics are encoded. As a future work, we aim to utilize our approach for building rehabilitation systems for ASC sufferers to provide formative feedback in performing certain facial expressions correctly. 

\section*{Acknowledgement}
The authors would like to thank Ezgi Dede for her support in conducting the user study.

{\small
\bibliographystyle{ieee_fullname}
\bibliography{AU_editing.bib}
}

\end{document}